\documentclass[sigconf]{acmart}
\author{Ruoxuan Li}
\orcid{0000-0002-1187-004X}
\affiliation{%
  \institution{Columbia University}
  \city{New York}
  \state{NY}
  \country{USA}
}
\email{rl3403@columbia.edu}

\author{Bruce Kogut}
\orcid{0000-0003-1355-9738}
\affiliation{%
  \department{Department of Sociology and Columbia Business School}
  \institution{Columbia University}
  \city{New York}
  \state{NY}
  \country{USA}
}
\email{bk2263@columbia.edu}

\AtBeginDocument{%
  }
\copyrightyear{2026}
\acmYear{2026}
\setcopyright{cc}
\setcctype{by}
\acmConference[CIKM '26]{Proceedings of the 35th ACM International Conference on Information and Knowledge Management}{November 7--11, 2026}{Rome, Italy}
\acmBooktitle{Proceedings of the 35th ACM International Conference on Information and Knowledge Management (CIKM '26), November 7--11, 2026, Rome, Italy}
\acmDOI{10.1145/3799682.3841048}
\acmISBN{979-8-4007-2539-5/2026/11}
\begin{document}
\title{Dynamic Topic Modeling for Cross-Corpus Temporal Analysis}
\begin{abstract}
Dynamic Embedded Topic Models (D-ETM) provide an interpretable framework for modeling temporal semantic evolution, but cross-corpus comparison remains difficult because topics are often learned independently and aligned only after training, a process that does not guarantee stable topic correspondence across corpora and time. To address this problem, we propose a D-ETM framework that first learns a common dynamic topic space over a merged multi-corpus collection, which we call the shared backbone, then introduces corpus-specific residual adaptation around the frozen backbone without creating separate latent topic spaces. This design preserves a shared topic index for cross-corpus comparison while allowing each corpus to specialize lexically. We evaluate the framework on three temporally structured corpora spanning 97 years: the Corpus of Historical American English, Harvard Business Review, and International Labour Review. Residual adaptation improves corpus-specific fit relative to the shared backbone while preserving the same-index cross-corpus topic trajectories, achieving substantially stronger alignment than full fine-tuning from the same backbone, with \(97.5 \pm 0.7\%\) versus \(17.9 \pm 1.1\% \) trajectory Retrieval@1, as well as stronger alignment than independent training with post-hoc Hungarian matching. These results suggest that incorporating topic alignment into the model can support more stable over-time cross-corpus comparisons while retaining corpus-specific lexical variation.

\end{abstract}

\renewcommand{\shortauthors}{Ruoxuan Li and Bruce Kogut}
\maketitle

\section{Introduction}
Computational analysis of large-scale text collections has become a well-established practice in social science, where topic models are widely used to discover and interpret latent thematic patterns in textual corpora \citep{grimmer2013text,taddy2013multinomial,evans2016machine,kozlowski2019geometry,quinn2010political,roberts2014structural}.  
Latent Dirichlet Allocation (LDA) provides a foundational probabilistic framework by representing documents as mixtures of latent topics, where each topic is a distribution over words \citep{blei2003latent}. For over-time text analysis, Dynamic Topic Models (DTM) extend this framework to sequential corpora by allowing topic-word distributions to evolve over time \citep{blei2006dynamic}. More recent neural topic models improve topic representations by using distributed word and topic embeddings. The Dynamic Embedded Topic Model (D-ETM) represents time-specific topic-word distributions through word and topic embeddings and learns smooth topic trajectories under a random-walk prior \citep{dieng2019dynamic}. These trajectories make D-ETM attractive for historical and social-science text analysis, where researchers often need interpretable themes that can be traced over long periods and related to social, political, or economic context \citep{grimmer2013text, roberts2014structural, roberts2019stm}.

Although D-ETM is well suited for modeling temporal semantic evolution within a single corpus, it introduces additional complexity in cross-corpus analysis. A common approach for comparing multiple corpora is to model each corpus separately and then align the resulting topics post hoc \citep{lu2019topic, bystrov2022cross, adam2025bidirectional}. However, post-hoc topic alignment is non-trivial: independently trained topic models do not guarantee that topic indices correspond across runs or corpora. Even two equivalent topic models can assign the same semantic topic to different indices, so topic comparison requires an explicit alignment procedure \citep{yang2016improving}. Prior work on topic model stability has therefore treated alignment as a prerequisite for comparing topic models, often relying on assignment-based procedures such as the Hungarian algorithm and similarity or divergence measures over topic-word distributions \citep{miller2017topic}. Such post-hoc alignment procedures become more difficult in over-time cross-corpus analysis using D-ETM, where the learned topic trajectories must remain identifiable across both corpora and time. A previous study on cross-corpus topic trends notes that such settings require identifying shared topics and their temporal evolution, while also acknowledging that there is no guarantee that matching topics can be found ex post \citep{bystrov2022cross}.

Given these challenges, we propose a shared-backbone D-ETM framework\footnote{Code is available at \url{https://github.com/TomlandHilarious/cross-corpus-dynamic-topic-modeling}.} for over-time cross-corpus analysis. Instead of training separate dynamic topic models and aligning the resulting topics after training, our proposed framework learns a common semantic coordinate system across corpora and time over the merged multi-corpus collection, which we define as the shared dynamic topic backbone. We then introduce source-specific residual adaptation around the shared backbone, allowing each corpus to specialize while preserving same-index topic correspondence across corpora. In this way, the framework treats cross-corpus comparability as a modeling assumption rather than as a post-hoc alignment problem. The contributions of our work are as follows:
\newline
\noindent\textbf{(1)}
We propose a shared-backbone D-ETM framework for over-time cross-corpus topic analysis, enabling corpus-specific adaptation while preserving aligned topic identities across corpora.

\noindent\textbf{(2)}
We evaluate the trade-off between source-specific fit and cross-corpus alignment under independent training, source-specific adaptation, and full fine-tuning.

\noindent\textbf{(3)}
We provide qualitative case studies showing how aligned topic trajectories support comparative over-time interpretation across heterogeneous corpora.

\section{Related Work}
\subsection{Dynamic Topic Modeling}
Dynamic topic modeling studies how latent themes evolve over time in temporally ordered document collections.\footnote{
Contextualized topic models and clustering-based approaches such as BERTopic are widely used for semantic topic discovery in applied text analysis \citep{devlin2019bert,bianchi2021pretraining,grootendorst2022bertopic}. They are outside the scope of this work because standard formulations are not designed around explicit temporal topic trajectories or source-specific drift parameters, which are central to our inference setting.
}
LDA provides the foundation for probabilistic topic modeling by representing each document as a mixture of latent topics, each of which is a distribution over vocabulary words \citep{blei2003latent}. However, LDA assumes a static topic set and cannot capture how topics shift over time. DTM addresses this limitation by allowing topic-word distributions to evolve across discrete time steps via a state-space model \citep{blei2006dynamic}. Subsequent work has explored alternative temporal representations. Topics over Time replaces discrete time slices with continuous timestamps to model topical trends \citep{wang2006topics}, while continuous-time DTM uses Brownian motion to capture topic evolution over sequentially ordered documents \citep{wang2008continuous}. Recent neural extensions improve topic representations and enable more scalable inference in dynamic topic models. D-ETM combines dynamic topic modeling with embedding-based topic representations. It parameterizes time-specific topic-word distributions through topic and word embeddings and encourages smooth topic trajectories through a random-walk prior over topic embeddings \citep{dieng2019dynamic}. This embedding-space view is also central to ETM, which represents topics and words in a shared vector space and helps topic models handle large and heavy-tailed vocabularies more effectively \citep{dieng2020topic}. More recent work further enriches dynamic topic modeling by capturing dependencies among co-evolving topics. For instance, DSNTM applies self-attention across topics to model branching and merging behaviors over time \citep{miyamoto2023dynamic}. Despite these advances, existing methods primarily focus on modeling topic evolution within a single temporally ordered corpus, not on preserving topic correspondence across multiple corpora.
\subsection{Cross-Corpus Topic Alignment}
Cross-corpus topic alignment aims to establish correspondences between topics learned from different document collections. One common strategy is to train topic models independently on each corpus and align the resulting topics in a post-hoc step. In cross-corpus topic trend analysis, for example, topics are first identified within each corpus, then matched across corpora, and finally compared through their temporal trajectories \citep{bystrov2022cross}. Bidirectional Topic Matching (BTM) adopts a similar post-hoc strategy, where it trains separate topic models on two corpora and applies each model to the other corpus to identify shared and corpus-specific topics \citep{adam2025bidirectional}. Other methods use topic-level similarity or divergence measures to compare topics across multiple corpora without joint training \citep{lu2019topic}. While these approaches make cross-corpus comparison possible, the resulting correspondences depend on the chosen topic representation, similarity measure, and matching criterion. This problem becomes more difficult in dynamic topic models, where alignment must hold not only between static topic-word distributions, but across evolving topic trajectories over time. 

Another line of work builds topic alignment directly into the modeling process. Polylingual topic models use comparable document tuples to learn topics that are aligned across languages \citep{mimno2009polylingual}. Subsequent multilingual topic models relax the requirement for closely parallel corpora by learning soft, weighted links between cross-lingual topics instead of enforcing strict one-to-one correspondence \citep{yang2019multilingual}. Cross-collection topic models take a similar approach, jointly learning shared and collection-specific topic structure across multiple document collections \citep{paul2009cross}. While these methods show that topic alignment can be incorporated into the modeling process, they are not designed for dynamic topic trajectories over a shared temporal span. Our work extends this modeling-oriented view of alignment to the D-ETM setting, where topic identity must remain comparable across both corpora and time.
\subsection{Topic Model Evaluation}
Topic model evaluation typically combines predictive and interpretable metrics. Predictive measures such as held-out likelihood and perplexity assess how well a model explains unseen documents \citep{wallach2009evaluation}, but better predictive fit does not necessarily yield more interpretable topics \citep{chang2009reading}. As a result, topic modeling work commonly reports semantic coherence measures such as normalized pointwise mutual information (NPMI), which evaluate whether high-probability topic words tend to co-occur more often than expected by chance \citep{newman2010automatic,mimno2011optimizing,lau2014machine}. In social-science applications, coherence is often considered alongside exclusivity, since coherent topics may still be dominated by common vocabulary and fail to distinguish one topic from another \citep{roberts2014structural,roberts2019stm}; neural topic models similarly report coherence, diversity, and topic quality to assess interpretability and redundancy \citep{dieng2020topic}. For dynamic topic models, evaluation must also account for temporal consistency, since year-wise topic quality can remain stable even when topic transitions are disrupted \citep{karakkaparambil-james-etal-2024-evaluating}.

However, these metrics assess topic quality after topic identities have already been defined within a model. They do not directly test whether topic indices provide comparable semantic coordinates across corpora. Independently trained models may produce coherent and temporally smooth topics within each corpus while still failing to preserve same-index topic correspondence across corpora and time. We therefore introduce trajectory-level alignment metrics to evaluate whether topic identities remain comparable in the cross-corpus dynamic setting.

\begin{figure*}[t]
\centering
\includegraphics[width=0.9\textwidth]{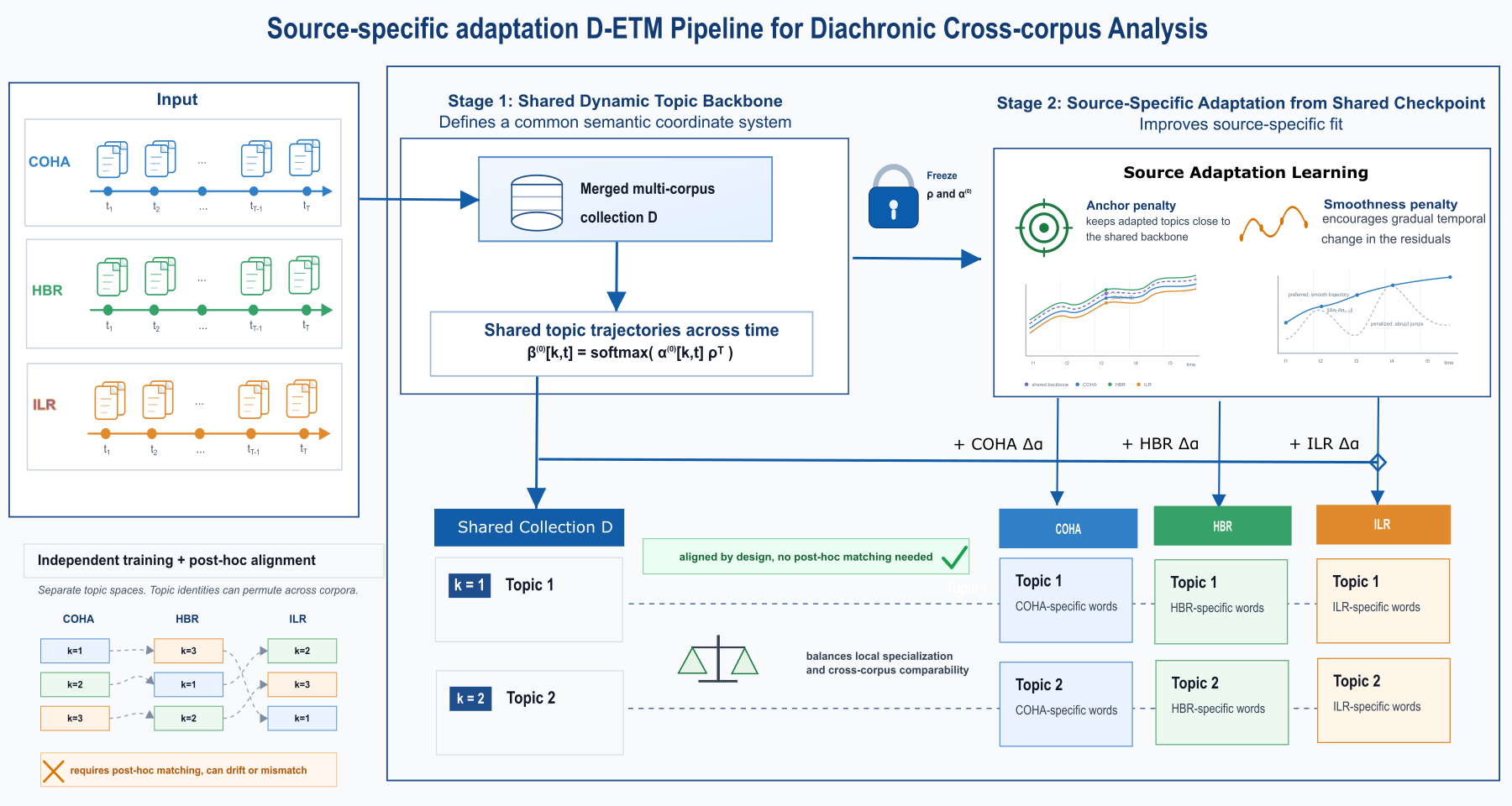}
\caption{Overview of the two-stage shared-backbone framework: merged-corpus backbone training followed by source-specific residual adaptation.}
\Description{Overview of the two-stage shared-backbone topic modeling framework with merged-corpus training followed by source-specific residual adaptation.}
\label{fig:method_framework}
\end{figure*}
\section{Preliminaries}
\subsection{Problem Definition}

Inspired by prior research on semantics in social science, we define the problem of dynamic topic modeling over multiple corpora that differ in domain and style but are aligned over a shared temporal span. Let $\mathcal{C}=\{c_1, c_2, \dots, c_S\}$ denote a collection of $S$ corpora, and let the shared temporal span be indexed by $t \in \{1, \dots, T\}$, where each document is associated with both a source label $s \in \{1, \dots, S\}$ and a time index $t$. Our goal is to model topic evolution over time while enabling reliable comparison of thematic structure across corpora, where an inherited topic index remains closer across sources than alternative topic indices after source-specific adaptation, as evaluated by the trajectory-alignment metrics in Section~\ref{sec:evaluation_metrics}.

For a fixed number of topics $K$, we seek to learn topic trajectories indexed by $k \in \{1,\dots,K\}$. In a standard single-corpus dynamic topic model, topic index $k$ identifies an evolving topic trajectory within one corpus. In the cross-corpus setting, however, we want topic identity to be preserved across both time and source, where topic $k$ in corpus $c_i$ and topic $k$ in corpus $c_j$ ideally refer to the same thematic context while allowing each topic to be expressed through source-specific vocabulary.

\subsection{Dynamic Embedded Topic Model}

The Dynamic Embedded Topic Model (D-ETM) represents topics in a continuous embedding space and models their evolution over ordered time slices \citep{dieng2019dynamic}. Let $V$ denote the vocabulary size, $K$ the number of topics, $L$ the embedding dimension, and $T$ the number of time slices. D-ETM maintains a word embedding matrix $\rho \in \mathbb{R}^{V \times L}$, where each row corresponds to the embedding of a vocabulary term. For each topic $k \in \{1,\dots,K\}$ and time slice $t \in \{1,\dots,T\}$, the model learns a time-specific topic embedding $\alpha_{k,t} \in \mathbb{R}^{L}$.

The sequence $\{\alpha_{k,t}\}_{t=1}^{T}$ defines the temporal trajectory of topic $k$. D-ETM encourages this trajectory to evolve smoothly by placing a random-walk prior over the topic embeddings:
\[
\alpha_{k,t} \mid \alpha_{k,t-1}
\sim
\mathcal{N}(\alpha_{k,t-1}, \sigma_\alpha^2 I),
\qquad t=2,\dots,T.
\]
Given $\alpha_{k,t}$ and the shared word embedding matrix $\rho$, the topic-word distribution for topic $k$ at time $t$ is
\[
\beta_{k,t}
=
\mathrm{softmax}(\alpha_{k,t}\rho^\top),
\]
where $\beta_{k,t} \in \Delta^{V-1}$ is a probability distribution over the vocabulary.

\section{Methodology}
\subsection{Proposed Framework Architecture}
Figure~\ref{fig:method_framework} summarizes the end-to-end structure of the framework, including Stage 1 backbone training and Stage 2 source adaptation.
\subsubsection{Shared Dynamic Topic Backbone}

In the first stage, we train a single D-ETM on the merged multi-corpus collection. 
Let
$\mathcal{D}=\{(x_d,s_d,t_d)\}_{d=1}^{N}$ denote the merged multi-corpus collection, where document
$x_d$ has source label $s_d \in \{1,\dots,S\}$ and time index
$t_d \in \{1,\dots,T\}$. The source label records which corpus the document comes from. Let $N_s$ denote the number of documents from source $s$. To prevent larger corpora
from dominating the shared backbone, we assign each document from source $s$ a
source-aware weight
\[
w_s = \frac{N}{S N_s}.
\]

This weighting gives each corpus equal total weight in the Stage 1 objective. We then
maximize the weighted D-ETM objective
\[
\mathcal{L}_{\mathrm{backbone}}
=
\sum_{d=1}^{N}
w_{s_d}
\mathcal{L}_{\mathrm{D\text{-}ETM}}(x_d,t_d),
\]
where $\mathcal{L}_{\mathrm{D\text{-}ETM}}(x_d,t_d)$ denotes the document-level D-ETM
evidence lower bound for document $x_d$ observed at time $t_d$.

Using the D-ETM formulation introduced above, Stage 1 learns a shared dynamic topic
trajectory for each topic:
\[
\alpha^{(0)}_{k,1:T}
=
\{\alpha^{(0)}_{k,1},\dots,\alpha^{(0)}_{k,T}\},
\]
where $\alpha^{(0)}_{k,t} \in \mathbb{R}^{L}$ represents the backbone embedding of
topic $k$ at time slice $t$. The superscript $(0)$ indicates that these are shared
backbone parameters, before any source-specific adaptation is introduced. As in
D-ETM, these topic embeddings evolve smoothly over time under the random-walk prior.

Given a shared word embedding matrix $\rho \in \mathbb{R}^{V \times L}$ over the merged
vocabulary, the corresponding backbone topic-word distribution is
\[
\beta^{(0)}_{k,t}
=
\mathrm{softmax}(\alpha^{(0)}_{k,t}\rho^\top),
\]
where $\beta^{(0)}_{k,t} \in \Delta^{V-1}$ gives a probability distribution over the
merged vocabulary. For a document $x_d$ observed at time $t_d$, the decoder uses the
time-specific backbone topics
$\{\beta^{(0)}_{1,t_d},\dots,\beta^{(0)}_{K,t_d}\}$.

The purpose of Stage 1 is to learn a common semantic coordinate system across the merged collection before any corpus-specific residual adaptation is introduced. Since all corpora share the same dynamic topic trajectories \(\alpha^{(0)}_{k,1:T}\), word embedding matrix \(\rho\), and backbone topic-word distributions \(\beta^{(0)}_{k,t}\), topic index \(k\) is defined with respect to a common dynamic backbone rather than separately within each corpus. This makes same-index comparison meaningful at the backbone level. Stage 2 then introduces source-specific residuals as controlled deviations from this shared topic coordinate.
\subsubsection{Source-Specific Adaptation}

To preserve cross-corpus comparability while allowing corpus-local specialization, Stage 2 models each source as a residual perturbation around the shared dynamic backbone. For each source $s \in \{1,\dots,S\}$, topic $k \in \{1,\dots,K\}$, and time slice $t \in \{1,\dots,T\}$, we introduce a source-specific residual topic offset $\Delta \alpha_{s,k,t} \in \mathbb{R}^{L}$. The source-adapted topic embedding is therefore defined as
\[
\tilde{\alpha}_{s,k,t}
=
\alpha^{(0)}_{k,t}
+
\Delta \alpha_{s,k,t},
\]
where $\alpha^{(0)}_{k,t}$ is the shared backbone topic embedding learned in Stage 1. Given the shared word embedding matrix $\rho$, the source-specific topic-word distribution is
\[
\beta^{(s)}_{k,t}
=
\mathrm{softmax}(\tilde{\alpha}_{s,k,t}\rho^\top)
=
\mathrm{softmax}\big((\alpha^{(0)}_{k,t}+\Delta \alpha_{s,k,t})\rho^\top\big).
\]
The residual-adaptation design is conceptually related to parameter-efficient adaptation, where a shared backbone is kept fixed while a small number of task- or domain-specific parameters are learned \citep{houlsby2019parameter,hu2022lora}. In our setting, the learnable residuals \(\Delta\alpha_{s,k,t}\) model corpus-specific drift around a shared D-ETM topic trajectory. This allows each corpus to adjust the lexical realization of a shared topic while preserving the common topic coordinate system needed for diachronic cross-corpus comparison. 

We regularize the residual offsets in two complementary ways. The anchor penalty controls the overall magnitude of source-specific drift by keeping \(\Delta\alpha_{s,k,t}\) close to the shared backbone:
\[
\mathcal{L}_{\mathrm{anchor}}
=
\lambda_{\mathrm{anchor}}
\sum_{s=1}^{S}\sum_{k=1}^{K}\sum_{t=1}^{T}
\left\|
\Delta \alpha_{s,k,t}
\right\|_2^2.
\]
A larger \(\lambda_{\mathrm{anchor}}\) therefore keeps the source-adapted topic embedding
\(\tilde{\alpha}_{s,k,t}\) closer to the shared trajectory
\(\alpha^{(0)}_{k,t}\), preserving cross-corpus topic identity more strongly.

The temporal smoothness penalty controls how abruptly the source-specific drift changes over time:
\[
\mathcal{L}_{\mathrm{smooth}}
=
\lambda_{\mathrm{smooth}}
\sum_{s=1}^{S}\sum_{k=1}^{K}\sum_{t=2}^{T}
\left\|
\Delta \alpha_{s,k,t} - \Delta \alpha_{s,k,t-1}
\right\|_2^2.
\]
A larger \(\lambda_{\mathrm{smooth}}\) discourages sharp changes in the residual offsets between adjacent time slices, encouraging corpus-specific specialization to evolve smoothly over the temporal trajectory.

During Stage 2, the shared backbone parameters \(\alpha^{(0)}_{k,t}\) and \(\rho\) are frozen, and the residuals \(\Delta\alpha_{s,k,t}\) are initialized at zero so that each source begins from the shared backbone with \(\tilde{\alpha}_{s,k,t}=\alpha^{(0)}_{k,t}\). For a document \(x_d\) from source \(s_d\) and time slice \(t_d\), reconstruction is evaluated using the source-specific topic-word distributions \(\{\beta^{(s_d)}_{k,t_d}\}_{k=1}^{K}\). Gradients from the reconstruction loss therefore update the corresponding residual offsets \(\Delta\alpha_{s_d,k,t_d}\) through the softmax parameterization, while the shared backbone remains fixed. The trainable parameters are the residual offsets and the amortized inference networks. We minimize an adaptation loss consisting of the reconstruction negative log-likelihood, a warmup-scheduled Kullback--Leibler (KL) divergence term that regularizes the inferred document-topic mixture \(q_\phi(\theta_d \mid x_d,\eta_{t_d})\) toward its prior \(p(\theta_d \mid \eta_{t_d})\), and the two residual regularization terms:
\[
\begin{aligned}
\mathcal{J}_{\mathrm{adapt}}
&=
\sum_{d=1}^{N}
 w_d
\Bigl[
-\mathbb{E}_{q_\phi(\theta_d \mid x_d, \eta_{t_d})}
\log p(x_d \mid \theta_d, \beta^{(s_d)}_{:,t_d})
\\
&\quad+
\omega_\theta(e)
\mathrm{KL}\big(q_\phi(\theta_d \mid x_d, \eta_{t_d}) \,\Vert\, p(\theta_d \mid \eta_{t_d})\big)
\Bigr]
\\
&\quad+
\mathcal{L}_{\mathrm{anchor}}
+
\mathcal{L}_{\mathrm{smooth}},
\end{aligned}
\]
where \(w_d = 1/(S|\mathcal{D}_{s_d}|)\) is a source-aware document weight, \(e\) denotes the adaptation epoch, and \(\omega_\theta(e)\) is a linear warmup schedule capped at \(\kappa_\theta\):
\[
\omega_\theta(e)
=
\min\left(\frac{e}{E_{\mathrm{warm}}},1\right)\kappa_\theta.
\]

\section{Experiments}
\subsection{Datasets}
We conduct experiments on three temporally structured corpora: the
Harvard Business Review (HBR) \citep{harvardBusinessReview},
the International Labour Review (ILR) \citep{internationalLabourReview},
and the Corpus of Historical American English (COHA)
\citep{davies2010coha}.
Prior social-science research has used HBR and ILR to compare business and labor discourse through static LDA \citep{disko2022efficient}. \footnote{Additional details on the original
retrieval and construction of the HBR and ILR corpora are documented
in an unpublished project memo~\citep{kogut2026sharedbackbone}.} Our study continues the use of these two datasets, while introducing COHA as a third, general-domain point of triangulation. Together, the three corpora cover overlapping historical periods but represent distinct linguistic domains: HBR captures managerial and business discourse, ILR captures labor, employment, and social-policy discourse, and COHA provides a broader baseline for historical American English. This combination allows us to test whether a shared dynamic topic space can capture common temporal structures while preserving corpus-specific specialization.

All corpora are restricted to the period 1922--2019 and grouped into
non-overlapping five-year bins, yielding $T=20$ time slices. To make COHA more
comparable to the periodical structure of HBR and ILR, we restrict COHA to its
\textsc{Magazine} and \textsc{News} genres. Since COHA remains much larger than the
other two corpora after this restriction, we downsample COHA chunks within each
year using a fixed sampling rate $\rho_{\mathrm{COHA}}=0.3$, bringing its effective corpus size close to HBR and ILR while preserving its temporal distribution. All documents are lowercased, cleaned, tokenized, and segmented into
non-overlapping chunks of at most 500 tokens. Each chunk inherits the source
label and publication year of its original document. We construct a shared
vocabulary from the training split of the merged corpus using a global
document-frequency filter with \texttt{min\_df}=100 and \texttt{max\_df}=0.6. These thresholds remove extremely rare and highly ubiquitous terms from the shared vocabulary and are held fixed across conditions. 
Documents with fewer than 50 post-vocabulary tokens are removed. The final
shared vocabulary contains 19{,}433 words. Train, validation, and test splits are created at the original-document level, so that chunks from the same document never appear
in different splits. We use stratified shuffling over the joint
$(\text{year}, \text{source})$ label and allocate 80\% of documents to training,
5\% to validation, and 15\% to test, as shown in Table~\ref{tab:dataset_stats}.

\begin{table}[t]
\centering
\caption{Dataset statistics after preprocessing. Counts refer to 500-token chunks in the merged-vocabulary configuration.}
\label{tab:dataset_stats}
\small
\begin{tabular}{l r r r r}
\toprule
Corpus & Train & Valid & Test & Total \\
\midrule
COHA (Mag.\,/\,News, $\rho{=}0.3$) & 32{,}460 & 1{,}999 & 6{,}042 & 40{,}501 \\
HBR                                & 33{,}513 & 1{,}777 & 6{,}358 & 41{,}648 \\
ILR                                & 32{,}615 & 2{,}169 & 5{,}821 & 40{,}605 \\
\midrule
\textbf{Merged D}                    & 98{,}588 & 5{,}945 & 18{,}221 & 122{,}754 \\
\bottomrule
\end{tabular}

\footnotesize
$T=20$ five-year bins covering 1922--2019; shared vocabulary $|V|=19{,}433$
with \texttt{min\_df}=100 and \texttt{max\_df}=0.6.
\end{table}
\subsection{Setup}
\label{sec:experimental_setup}
We evaluate five training conditions designed to separate the effects of vocabulary choice, joint multi-corpus training, residual adaptation, and full fine-tuning, summarized in Table~\ref{tab:training_conditions}. 

\begin{table}[t]
\centering
\caption{Training and comparison conditions.}
\label{tab:training_conditions}
\small
\begin{tabular}{l p{0.72\linewidth}}
\toprule
\textbf{Condition} & \textbf{Description} \\
\midrule
\textbf{Ind-CS} 
& Independently train one D-ETM per corpus using that corpus's own active vocabulary. This represents the standard independent-training setting, where cross-corpus comparison requires post-hoc topic matching. \\

\textbf{Ind-MV} 
& Independently train one D-ETM per corpus using the shared merged vocabulary. This controls for vocabulary differences while keeping the training process corpus-specific. \\

\textbf{SB-Joint} 
& Train a single D-ETM on the merged multi-corpus collection. This model learns the shared dynamic topic backbone used to define common topic indices across corpora. \\

\textbf{SB-RA} 
& Initialize from \textbf{SB-Joint}, freeze the shared backbone, and learn source-specific residual topic offsets. This is the proposed method for corpus-local lexical specialization while preserving shared topic identities. \\

\textbf{SB-FT} 
& Initialize from \textbf{SB-Joint} and fully fine-tune all model parameters separately on each corpus. This provides a less constrained adaptation baseline that tests what happens when the shared backbone is not preserved. \\
\bottomrule
\end{tabular}
\end{table}
\subsubsection{Training protocol.}
The Stage 1 conditions, \textbf{Ind-CS}, \textbf{Ind-MV}, and \textbf{SB-Joint}, use the same core D-ETM configuration unless otherwise noted. We set the number of topics to \(K=20\) and use non-overlapping five-year temporal bins as a stable common operating point for comparing training regimes across corpora. This choice is driven by our goal  to evaluate whether topics learned across the three corpora remain comparable under the same modeling conditions. In preliminary experiments with independently trained D-ETM models on our corpora, larger topic counts produced more repeated or weakly interpretable topics, especially for HBR and ILR. Finer temporal bins also made trajectories less stable because some source-year slices contain sparse or uneven data. Since the goal of this paper is to evaluate built-in alignment of dynamic topic trajectories across corpora, we fix both topic count and temporal granularity across all models. We use \(K=20\) and five-year bins as a practical common setting that balances interpretability and cross-corpu.

Optimization uses learning rate $5 \times 10^{-5}$ for 80 epochs. We set the D-ETM transition variance to $\delta=0.01$, use $\texttt{KL\_ALPHA\_SCALE}=10^{-6}$ to downweight the KL term associated with the random-walk prior over dynamic topic embeddings, and apply a 50-epoch KL warmup with maximum weight $\texttt{KL\_WEIGHT\_MAX}=0.9$. These KL settings were chosen empirically because, without sufficient downweighting and warmup, the trajectory KL term dominated the reconstruction objective early in training and limited the model's ability to learn coherent topic-word structure. \textbf{Ind-CS} and \textbf{Ind-MV} are independently trained corpus-specific baselines, while \textbf{SB-Joint} is trained on the merged multi-corpus collection and provides the shared-backbone checkpoint used in Stage 2.

The Stage 2 conditions, \textbf{SB-RA} and \textbf{SB-FT}, are initialized from the \textbf{SB-Joint} checkpoint. In \textbf{SB-RA}, the shared word embedding matrix $\rho$ and shared backbone topic embeddings $\alpha^{(0)}_{k,t}$ are frozen, while the source-specific residuals $\Delta\alpha_{s,k,t}$ and inference networks remain trainable. The residuals are initialized at zero and optimized for 20 adaptation epochs with learning rate $1 \times 10^{-5}$. During adaptation, we retain only the document-topic KL term and use a shorter warmup schedule, with $\texttt{ADAPT\_WARMUP\_EPOCHS}=5$ and maximum KL weight $\texttt{ADAPT\_KL\_THETA\_MAX}=0.3$.

\subsubsection{Post-hoc Topic Matching Baseline.}
For independently trained models, topic indices are arbitrary and cannot be compared directly across corpora. We therefore use post-hoc topic matching as the alignment baseline for \textbf{Ind-CS} and \textbf{Ind-MV}, following prior cross-corpus topic comparison work \citep{bystrov2022cross}. To match our trajectory-level alignment evaluation, we perform post-hoc matching over full topic trajectories rather than separately within each time slice.

For source \(s\) and topic \(k\), we represent the full topic trajectory as a probability distribution over word--time pairs:
\[
P_{s,k}(w,t)
=
\frac{1}{T}\beta^{(s)}_{k,t}(w).
\]
For each corpus pair \((s,s')\), we then construct a trajectory-level cost matrix using Jensen--Shannon divergence (JSD), computed with base-2 logarithms \citep{lin1991divergence}:
\[
C^{(s,s')}_{k,\ell}
=
\mathrm{JSD}
\left(
P_{s,k},
P_{s',\ell}
\right).
\]
Finally, we apply the Hungarian algorithm to find the one-to-one topic matching that minimizes total trajectory divergence \citep{kuhn1955hungarian}:
\[
\pi^{\star}
=
\arg\min_{\pi}
\sum_{k=1}^{K}
C^{(s,s')}_{k,\pi(k)}.
\]

For \textbf{Ind-MV}, trajectory JSD is computed directly in its vocabulary space. For \textbf{Ind-CS}, we restrict topic-word distributions to the vocabulary intersection and renormalize them before computing JSD, giving the independently trained models a common support for post-hoc comparison.

\subsection{Evaluation Metrics}
\label{sec:evaluation_metrics}
We evaluate all training conditions along three dimensions: predictive fit, full-reference topic
quality, and cross-corpus trajectory alignment. Predictive fit and topic quality
follow standard topic modeling and D-ETM evaluation practice, while the
trajectory alignment metrics are designed for the cross-corpus dynamic setting.

\subsubsection{Predictive Fit and Topic Quality}
We report held-out perplexity on the shared merged test split for all five
training conditions, following standard
topic modeling evaluation practice \citep{blei2003latent, blei2006dynamic,
dieng2019dynamic}. We use perplexity mainly to assess whether the proposed adaptation remains in a comparable predictive range. Following the D-ETM evaluation protocol \citep{dieng2019dynamic}, we also report topic diversity (TD), measured as the proportion of unique words among the top topic words, pairwise NPMI-based topic coherence (TC), computed over high-probability topic words with \(-1\) assigned to word pairs with zero document co-occurrence, and topic quality (TQ), computed as \(\mathrm{TD}\times\mathrm{TC}\).
\subsubsection{Cross-Corpus Trajectory Alignment}

Our main alignment metrics compare entire topic trajectories across corpora.
For source $s$ and topic $k$, we define a trajectory distribution over
word--time pairs:
\[
P_{s,k}(w,t)
=
\frac{1}{T}\beta^{(s)}_{k,t}(w).
\]
This distribution gives equal weight to each time bin and forms a probability
vector over the $T \times V$ word--time space. For each source pair $(s,s')$, we construct a trajectory-level JSD matrix:
\[
M^{(s,s')}_{k\ell}
=
\mathrm{JSD}
\left(
P_{s,k},
P_{s',\ell}
\right),
\]
Each entry measures the distance between the full trajectory of topic \(k\) in source \(s\) and topic \(\ell\) in source \(s'\). Based on this trajectory-level JSD matrix, 
we define four complementary trajectory-level alignment metrics based on \(M^{(s,s')}\). It is worth noting that, since all four metrics are derived from the same trajectory-level JSD matrix \(M^{(s,s')}\), they should not be interpreted as statistically independent measures of alignment quality. Instead, they capture complementary aspects of cross-corpus trajectory alignment.

\noindent\textbf{1) Same-index trajectory JSD.}
This metric measures the average trajectory distance between topics with the same inherited index. Lower values indicate that same-index topic trajectories remain closer across sources:
\[
\mathrm{SameJSD}_{\mathrm{traj}}(s,s')
=
\frac{1}{K}
\sum_{k=1}^{K}
M^{(s,s')}_{kk}.
\]

\noindent\textbf{2) Trajectory margin.}
This metric measures whether each same-index trajectory is closer than its nearest wrong-index alternative. Higher positive values indicate stronger separation between the same-index trajectory and the nearest wrong-index competitor:
\[
\mathrm{Margin}_{\mathrm{traj}}(s,s')
=
\frac{1}{K}
\sum_{k=1}^{K}
\left[
\min_{\ell \neq k} M^{(s,s')}_{k\ell}
-
M^{(s,s')}_{kk}
\right].
\]

\noindent\textbf{3) Trajectory Retrieval@1.}
This metric measures how often the nearest topic trajectory in the other source has the same index. Higher values indicate that the inherited topic index remains a more reliable cross-corpus semantic coordinate:
\[
\mathrm{R@1}_{\mathrm{traj}}(s,s')
=
\frac{1}{K}
\sum_{k=1}^{K}
\mathbb{1}
\left[
k
=
\arg\min_{\ell}
M^{(s,s')}_{k\ell}
\right].
\]

\noindent\textbf{4) Hungarian-matched trajectory JSD.}
This metric reports the minimum post-hoc one-to-one matching cost under the same trajectory JSD matrix \(M^{(s,s')}\). Lower values indicate that a lower-cost post-hoc alignment can be found, but this metric does not test whether the original shared topic index is preserved:
\[
\mathrm{HungJSD}_{\mathrm{traj}}(s,s')
=
\frac{1}{K}
\sum_{k=1}^{K}
M^{(s,s')}_{k,\pi^{\star}(k)},
\]
where \(\pi^\star\) is the minimum-cost one-to-one topic matching found by the Hungarian algorithm. 
\begin{table*}[t]
\centering
\small
\caption{Held-out perplexity across experimental conditions on the shared merged
test split. Lower values are better. Each row is evaluated on the same held-out
documents within the corresponding source.}
\label{tab:ppl_shared_split}
\begin{tabular*}{\textwidth}{@{\extracolsep{\fill}}lrrrrrr}
\toprule
Source & $n_{\mathrm{test}}$ & Ind-CS & Ind-MV & SB-Joint & SB-RA & SB-FT \\
\midrule
COHA & 6,042 & 6,446.75 & 6,458.27 & 9,819.13 & 7,896.94 & \textbf{6,364.36} \\
HBR  & 6,358 & 2,479.74 & 2,530.01 & 2,570.88 & 2,396.78 & \textbf{2,127.39} \\
ILR  & 5,821 & 1,722.68 & 1,701.09 & 1,561.29 & 1,486.90 & \textbf{1,320.03} \\
\bottomrule
\end{tabular*}
\end{table*}
\subsection{Experiment Analysis}
\subsubsection{Comparison of Predictive Fit and Topic Quality}
\label{sec:qualitative_fit}
We first evaluate whether the shared-backbone framework provides reasonable
predictive fit and topic quality before examining cross-corpus alignment.
Table~\ref{tab:ppl_shared_split} reports held-out perplexity across all
experimental conditions on the shared merged test split. SB-FT achieves the
lowest perplexity on all three sources. On COHA, Ind-CS and Ind-MV obtain perplexities both substantially lower than SB-RA, indicating a predictive-fit cost for constrained residual adaptation on this corpus. On HBR and ILR, however, SB-RA improves over both independently trained
baselines. Across
all three sources, SB-RA also improves over SB-Joint. These results suggest
that residual adaptation recovers meaningful source-specific fit from the
shared backbone and remains broadly comparable to independently trained models,
although full fine-tuning remains strongest in predictive performance.

\begin{table}[t]
\centering
\small
\setlength{\tabcolsep}{5pt}
\renewcommand{\arraystretch}{1.08}
\caption{
Full-reference topic quality across experimental conditions.
}
\label{tab:topic_quality}
\begin{tabular}{lrrrr}
\toprule
\textbf{Setup} & \textbf{TD} & \textbf{TC} & \(\boldsymbol{\mathrm{C}_V}\) & \textbf{TQ} \\
\midrule
Ind-CS-COHA & 0.5843 &  0.0551 & 0.2300 &  0.0322 \\
Ind-CS-HBR  & 0.5304 &  0.0460 & 0.1968 &  0.0244 \\
Ind-CS-ILR  & 0.4494 &  0.0350 & 0.2172 &  0.0157 \\
\midrule
Ind-MV-COHA & 0.5810 &  0.0558 & 0.2229 &  0.0324 \\
Ind-MV-HBR  & 0.5015 &  0.0413 & 0.1975 &  0.0207 \\
Ind-MV-ILR  & 0.4824 &  0.0377 & 0.2198 &  0.0182 \\
\midrule
SB-Joint-ALL & 0.4538 &  0.1685 & 0.3041 &  0.0765 \\
\midrule
SB-RA-COHA  & 0.4611 &  0.0946 & 0.2536 &  0.0436 \\
SB-RA-HBR   & 0.4466 &  0.0638 & 0.1991 &  0.0285 \\
SB-RA-ILR   & 0.4335 & -0.0157 & 0.2240 & -0.0068 \\
\midrule
SB-FT-COHA  & 0.5690 &  0.1174 & 0.3027 &  0.0668 \\
SB-FT-HBR   & 0.5660 &  0.0979 & 0.2357 &  0.0554 \\
SB-FT-ILR   & 0.5059 &  0.0780 & 0.2353 &  0.0394 \\
\bottomrule
\end{tabular}
\end{table}
Table~\ref{tab:topic_quality} reports full-reference topic quality. SB-Joint obtains the highest NPMI-based coherence when evaluated against the merged reference corpus. Among source-specific models, SB-FT achieves the strongest topic quality across the three sources, consistent with its stronger source-specific predictive fit. SB-RA improves NPMI-based TC and TQ over
both independent baselines on COHA and HBR, but yields negative
NPMI-based coherence on ILR.
  
We therefore include \(\mathrm{C}_V\) as an additional diagnostic coherence measure \citep{roder2015exploring}. Unlike the pairwise NPMI-based TC used in the D-ETM protocol, \(\mathrm{C}_V\) uses a sliding-window co-occurrence representation and an indirect confirmation measure, making it less directly driven by individual zero co-occurrence word pairs. The \(\mathrm{C}_V\) results provide a more nuanced interpretation of the ILR case: although SB-RA has negative NPMI-based coherence on ILR, its \(\mathrm{C}_V\) coherence remains within the range of the baseline models and is higher than both independent ILR baselines. This suggests that the negative ILR NPMI reflects the sensitivity of pairwise co-occurrence-based TC to sparse word co-occurrence patterns in this corpus, and does not by itself indicate a loss of interpretable topic structure. Overall, SB-RA preserves usable topic quality relative to the independently trained baselines, even though it does not maximize coherence in all source-specific settings.

\subsubsection{Cross-Corpus Trajectory Alignment and Ablations}

Table~\ref{tab:source_to_source_alignment} reports trajectory-level cross-corpus alignment. For independently trained baselines, topic indices are arbitrary, so same-index metrics are not meaningful. We therefore report only post-hoc Hungarian trajectory JSD for \textbf{Ind-CS} and \textbf{Ind-MV}. Both independent baselines yield substantially higher post-hoc matching costs than SB-RA, indicating that independent training followed by post-hoc matching does not recover strong cross-corpus trajectory correspondence. 

For shared-backbone-initialized models, same-index metrics test whether the topic index inherited from \textbf{SB-Joint} remains meaningful after adaptation. The main \textbf{SB-RA} setting achieves the strongest alignment, with
same-index and Hungarian-matched trajectory JSD both at
\(0.169 \pm 0.001\), a positive margin of
\(+0.166 \pm 0.002\), and Retrieval@1 of \(97.5 \pm 0.7\%\),
showing that the inherited topic indices already realize the optimal
one-to-one matching.

The ablations show that alignment preservation is driven primarily by
the anchor penalty, with anchor-only adaptation performing nearly
identically to the full SB-RA model. Temporal smoothness also
contributes to alignment preservation, but its effect is weaker and
less stable across seeds. Removing both regularizers yields the lowest
Retrieval@1 among the SB-RA variants (\(33.8 \pm 4.3\%\)), yet still
outperforms SB-FT (\(17.9 \pm 1.1\%\)). More broadly, SB-FT achieves
substantially weaker inherited topic correspondence than the main
SB-RA setting across all alignment metrics, despite its stronger
predictive fit and topic quality. This suggests that both the explicit
regularization and the fixed-backbone residual parameterization
contribute to preserving the shared topic coordinate system.

\begin{table*}[t]
\centering
\caption{
Trajectory-level cross-corpus alignment across experimental and ablation conditions.
All results are averaged over the three corpus pairs: COHA--HBR, COHA--ILR, and HBR--ILR.
Shared-backbone-initialized adaptation conditions (SB-RA and SB-FT) are reported as
mean $\pm$ standard deviation over four Stage-2 random seeds using the same pretrained
SB-Joint backbone. Independently trained baselines are single runs and are evaluated
using post-hoc Hungarian matching.
}
\label{tab:source_to_source_alignment}

\small
\setlength{\tabcolsep}{4pt}
\renewcommand{\arraystretch}{1.08}

\begin{tabular*}{\textwidth}{@{\extracolsep{\fill}}lccrrrr}
\toprule
\textbf{Method}
& \(\boldsymbol{\lambda}_{\textbf{anchor}}\)
& \(\boldsymbol{\lambda}_{\textbf{smooth}}\)
& \shortstack{\textbf{Same}\\\textbf{JSD} \(\downarrow\)}
& \shortstack{\textbf{Hung.}\\\textbf{JSD} \(\downarrow\)}
& \textbf{Margin} \(\uparrow\)
& \textbf{R@1} \(\uparrow\) \\
\midrule

Ind-CS
& -- & --
& -- & 0.594 & -- & -- \\

Ind-MV
& -- & --
& -- & 0.619 & -- & -- \\

\midrule

\textbf{SB-RA}
& \(10^{-3}\)
& \(10^{-3}\)
& \(\mathbf{0.169 \pm 0.001}\)
& \(\mathbf{0.169 \pm 0.001}\)
& \(\mathbf{+0.166 \pm 0.002}\)
& \(\mathbf{97.5 \pm 0.7\%}\) \\

SB-RA
& \(3{\times}10^{-4}\)
& \(3{\times}10^{-4}\)
& \(0.334 \pm 0.004\)
& \(0.334 \pm 0.004\)
& \(+0.029 \pm 0.002\)
& \(70.4 \pm 0.8\%\) \\

SB-RA
& \(10^{-4}\)
& \(10^{-4}\)
& \(0.428 \pm 0.009\)
& \(0.424 \pm 0.008\)
& \(-0.034 \pm 0.005\)
& \(47.1 \pm 3.1\%\) \\

SB-RA
& \(10^{-3}\)
& \(0\)
& \(0.176 \pm 0.001\)
& \(0.176 \pm 0.001\)
& \(+0.155 \pm 0.001\)
& \(96.9 \pm 0.4\%\) \\

SB-RA
& \(0\)
& \(10^{-3}\)
& \(0.424 \pm 0.039\)
& \(0.422 \pm 0.038\)
& \(-0.019 \pm 0.023\)
& \(55.2 \pm 15.6\%\) \\

SB-RA
& \(0\)
& \(0\)
& \(0.478 \pm 0.017\)
& \(0.472 \pm 0.016\)
& \(-0.060 \pm 0.009\)
& \(33.8 \pm 4.3\%\) \\

\midrule

SB-FT
& -- & --
& \(0.586 \pm 0.006\)
& \(0.568 \pm 0.004\)
& \(-0.077 \pm 0.006\)
& \(17.9 \pm 1.1\%\) \\

\bottomrule
\end{tabular*}
\end{table*}

\begin{table}[t]
\centering
\caption{
Sensitivity of trajectory alignment to the number of topics.
All models use the same random seed within each topic-count setting.
Results are averaged over the three corpus pairs.
}
\label{tab:k_sensitivity}
\small
\setlength{\tabcolsep}{4pt}
\begin{tabular}{lrrrrr}
\toprule
$K$ & Method &
\shortstack{Same\\JSD $\downarrow$} &
\shortstack{Hung.\\JSD $\downarrow$} &
Margin $\uparrow$ &
R@1 $\uparrow$ \\
\midrule
10 & SB-RA & 0.257 & 0.257 & +0.106 & 85.0\% \\
   & SB-FT & 0.533 & 0.526 & -0.058 & 26.7\% \\
\midrule
20 & SB-RA & 0.170 & 0.170 & +0.164 & 98.3\% \\
   & SB-FT & 0.592 & 0.573 & -0.074 & 17.5\% \\
\midrule
30 & SB-RA & 0.106 & 0.106 & +0.196 & 100.0\% \\
   & SB-FT & 0.570 & 0.553 & -0.087 & 13.9\% \\
\bottomrule
\end{tabular}
\end{table}
\subsubsection{K-Sensitivity Test and Runtime Analysis}
We additionally evaluate sensitivity to the number of topics using
$K\in\{10,20,30\}$ while holding the random seed fixed.
As shown in Table~\ref{tab:k_sensitivity}, the alignment advantage
of SB-RA over SB-FT is preserved across all three topic counts.
SB-RA maintains substantially lower same-index trajectory JSD and
positive trajectory margins at each $K$, while SB-FT yields negative
margins throughout. Retrieval@1 similarly remains substantially
higher for SB-RA compared to SB-FT. These results indicate that the relative
alignment advantage of residual adaptation is not specific to the
default $K=20$ configuration.

We further characterize computational cost using the $K=10$ and
$K=30$ sensitivity runs, both executed on NVIDIA A100-SXM4-80GB
GPUs.\footnote{The original $K=20$ experiments were conducted on
NVIDIA A40 GPUs before runtime instrumentation was added, so we
restrict the runtime comparison to the two hardware-consistent A100
runs. Future work can evaluate runtime and memory scaling across a broader
range of topic counts and hardware configurations.} 
As shown in Table~\ref{tab:runtime}, increasing the topic count from
$K=10$ to $K=30$ increases SB-Joint training time and SB-RA adaptation time, while
peak allocated GPU memory also increases. 

\begin{table}[t]
\centering
\caption{Runtime and peak GPU memory for the $K$-sensitivity experiments on NVIDIA A100-SXM4-80GB GPUs. SB-FT runtime is summed across COHA, HBR, and ILR.}
\label{tab:runtime}
\small

\begin{tabular}{crrrr}
\toprule
$K$ &
\shortstack{SB-Joint\\time} &
\shortstack{SB-RA\\time} &
\shortstack{SB-FT\\time} &
\shortstack{Peak GPU\\memory} \\
\midrule
10 & 57m 26s & 13m 11s & 14m 53s & 3.14 GB \\
30 & 82m 13s & 18m 21s & 20m 07s & 7.51 GB \\
\bottomrule
\end{tabular}
\end{table}
\subsection{Case Study}
The quantitative results show that \textbf{SB-RA} preserves same-index topic trajectories across sources while allowing source-specific adaptation. This preservation is important not only for alignment metrics, but also for qualitative interpretation. In a standard D-ETM, the random-walk prior encourages each topic to evolve as a smooth time-indexed trajectory, and the learned topic-word distributions provide an observable way to inspect this temporal semantic change. Our framework extends this qualitative use of D-ETM to the cross-corpus setting. The shared backbone lets us examine whether the merged corpus learns historically meaningful base topics and broad temporal trends, while residual adaptation lets us compare how each corpus expresses the same shared topic trajectory through source-specific vocabulary. The case study therefore focuses on two questions. First, whether the shared backbone learns interpretable historical topic trajectories. Second, whether the source-adapted topics reveal meaningful corpus-local lexical specialization around the same topic coordinate.

 We use a simple Top-30 overlap check to summarize how closely each source-specific topic stays to the shared backbone (see Table~\ref{tab:top30_overlap}). For each source, topic, and time bin, we compare the source-specific Top-30 words with the shared-backbone Top-30 words. The overlap is the proportion of shared words between the two lists. We also report the average number of source-specific Top-30 words that do not appear in the shared Top-30 list. This gives a simple descriptive view of how much additional corpus-local vocabulary appears around the shared topic coordinate.  COHA has the highest average overlap with the shared backbone. This pattern may reflect COHA's role as a broad general-domain corpus. It is also possible that the shared backbone is learned from the merged collection and can therefore be more strongly shaped by the corpus with broader lexical coverage. 

\begin{table}[t]
\centering
\small
\caption{Average Top-30 vocabulary overlap with the shared backbone across all topics and time bins. }
\label{tab:top30_overlap}
\begin{tabular}{lcc}
\toprule
 & Avg. Top-30 overlap & Avg. \# outside shared Top-30 \\
\midrule
COHA & 0.853 $\pm$ 0.106 & 4.42 $\pm$ 3.19 \\
HBR  & 0.790 $\pm$ 0.103 & 6.31 $\pm$ 3.08 \\
ILR  & 0.791 $\pm$ 0.131 & 6.26 $\pm$ 3.93 \\
\bottomrule
\end{tabular}
\end{table}
\begin{table*}[t]
\centering
\scriptsize
\setlength{\tabcolsep}{2pt}
\renewcommand{\arraystretch}{1.08}

\caption{
Corpus-local Top-30 words for Topic 0. Source-local words are obtained by removing the corresponding shared-backbone Top-30 words from each source-specific Top-30 list.
}
\label{tab:topic0_source_local}

\newcommand{\topiccell}[1]{%
  \parbox[t]{0.205\textwidth}{\raggedright #1}%
}

\begin{tabular*}{\textwidth}{@{\extracolsep{\fill}}lcccc}
\toprule
&
\topiccell{\textbf{Industrial crisis}}
&
\topiccell{\textbf{Corporate management}}
&
\topiccell{\textbf{Workplace organization}}
&
\topiccell{\textbf{Digital services}}
\\
&
\topiccell{\textbf{1922--1942}}
&
\topiccell{\textbf{1947--1977}}
&
\topiccell{\textbf{1982--1992}}
&
\topiccell{\textbf{1997--2017}}
\\
\midrule

COHA
& \topiccell{man, credit, public, power, department, material, period, place}
& \topiccell{control, policy, operation, production, price, service, major, small}
& \topiccell{information, change, state, price, rate, power, group, small}
& \topiccell{people, world, group, price, return, board, industry, cost}
\\

HBR
& \topiccell{bank, rate, export, cost, material, credit, income, investment}
& \topiccell{profit, sale, capital, investment, product, price, rate, tax}
& \topiccell{price, capital, competition, investment, product, company, profit, corporate}
& \topiccell{cost, product, customer, sale, consumer, manager, stock, revenue}
\\

ILR
& \topiccell{wage, labor, worker, condition, factory, rate, income, country}
& \topiccell{worker, labor, country, wage, policy, income, estimate, level}
& \topiccell{policy, level, worker, labor, trade, economic, rate, price}
& \topiccell{social, worker, labor, job, pay, country, level, cost}
\\

\bottomrule
\end{tabular*}
\end{table*}

\begin{table}[t]
\centering
\scriptsize
\setlength{\tabcolsep}{0pt}
\renewcommand{\arraystretch}{0.9}
\setlength{\fboxsep}{2pt}
\setlength{\fboxrule}{0.4pt}

\caption{
Phase-level summaries for three topics selected from the trained shared-backbone model. 
}

\label{tab:multi_topic_shared_phases}

\newcommand{\topicbox}[1]{%
\fbox{\begin{minipage}[t][0.82in][t]{0.145\columnwidth}
\centering
#1
\end{minipage}}%
}

\newcommand{\phasehead}[1]{%
\begin{minipage}[t]{0.145\columnwidth}
\centering
#1
\end{minipage}%
}

\begin{tabular}{@{}c@{\hspace{0.095\columnwidth}}c@{\hspace{0.095\columnwidth}}c@{\hspace{0.095\columnwidth}}c@{}}

\phasehead{
{\bfseries Industrial}\\
{\bfseries crisis}\\
{\bfseries 1922--1942}
}
&
\phasehead{
{\bfseries Corporate /}\\
{\bfseries welfare}\\
{\bfseries 1947--1977}
}
&
\phasehead{
{\bfseries Workplace}\\
{\bfseries reorganization}\\
{\bfseries 1982--1992}
}
&
\phasehead{
{\bfseries Digital /}\\
{\bfseries labor market}\\
{\bfseries 1997--2017}
}
\end{tabular}

\textbf{Topic 0: Macroeconomic transformation}\\
\begin{tabular}{
@{}c
@{\hspace{2pt}\raisebox{-0.32in}{\Large$\Longrightarrow$}\hspace{2pt}}
c
@{\hspace{2pt}\raisebox{-0.32in}{\Large$\Longrightarrow$}\hspace{2pt}}
c
@{\hspace{2pt}\raisebox{-0.32in}{\Large$\Longrightarrow$}\hspace{2pt}}
c@{}
}
\topicbox{
industry\\
production\\
trade\\
credit\\
depression\\
economic\\
price\\
postwar
}
&
\topicbox{
company\\
management\\
market\\
product\\
capital\\
investment\\
cost\\
service
}
&
\topicbox{
employee\\
work\\
job\\
system\\
management\\
service\\
market\\
company
}
&
\topicbox{
internet\\
software\\
web\\
customer\\
market\\
service\\
platform\\
innovation
}
\end{tabular}

\textbf{Topic 15: Corporate management and business organization}\\
\begin{tabular}{
@{}c
@{\hspace{2pt}\raisebox{-0.32in}{\Large$\Longrightarrow$}\hspace{2pt}}
c
@{\hspace{2pt}\raisebox{-0.32in}{\Large$\Longrightarrow$}\hspace{2pt}}
c
@{\hspace{2pt}\raisebox{-0.32in}{\Large$\Longrightarrow$}\hspace{2pt}}
c@{}
}
\topicbox{
company\\
stock\\
cost\\
profit\\
price\\
market\\
sale\\
corporation
}
&
\topicbox{
company\\
management\\
cost\\
manager\\
product\\
executive\\
system\\
market
}
&
\topicbox{
company\\
manager\\
product\\
customer\\
technology\\
quality\\
strategy\\
process
}
&
\topicbox{
company\\
customer\\
leader\\
team\\
innovation\\
strategy\\
performance\\
organization
}
\end{tabular}

\textbf{Topic 16: Labor welfare and employment protection}\\
\begin{tabular}{
@{}c
@{\hspace{2pt}\raisebox{-0.32in}{\Large$\Longrightarrow$}\hspace{2pt}}
c
@{\hspace{2pt}\raisebox{-0.32in}{\Large$\Longrightarrow$}\hspace{2pt}}
c
@{\hspace{2pt}\raisebox{-0.32in}{\Large$\Longrightarrow$}\hspace{2pt}}
c@{}
}
\topicbox{
worker\\
employer\\
labor\\
wage\\
hour\\
insurance\\
benefit\\
woman
}
&
\topicbox{
worker\\
employment\\
labor\\
wage\\
social\\
benefit\\
union\\
pension
}
&
\topicbox{
employment\\
worker\\
labor\\
woman\\
employer\\
benefit\\
union\\
contract
}
&
\topicbox{
worker\\
labor\\
wage\\
employment\\
minimum\\
child\\
security\\
care
}
\end{tabular}
\end{table}

\begin{figure*}[t]
    \centering
    \includegraphics[width=\textwidth]{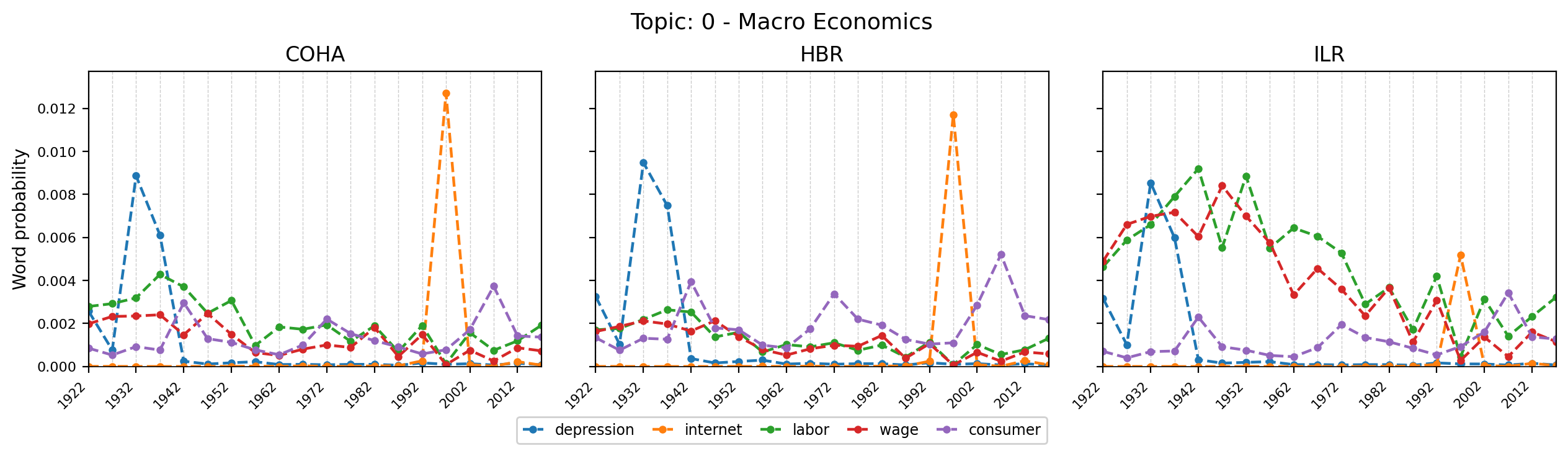}
    
    \caption{
    Raw topic-word probability trajectories for selected Topic 0 words across COHA, HBR, and ILR. The plotted values are source-specific topic-word probabilities over five-year bins. 
    }
    \Description{Line plots showing source-specific word-probability trajectories for selected words in aligned Topic 0 across COHA, HBR, and ILR over five-year time bins.}
    \label{fig:raw_beta_topic0}
\end{figure*}
\subsubsection{Historical Interpretability.}
We present three topic trajectories via the shared-backbone learning phase in Table~\ref{tab:multi_topic_shared_phases}. The four historical phases are derived from the trajectory of Topic 0 and are used as a common reference frame for presenting how these topics evolve over the same time period. This overview suggests that the D-ETM backbone learns meaningful shared topic trajectories on the merged corpus. Topic 0 captures a broad macroeconomic trajectory that is relevant across all three corpora, while Topic 15 and Topic 16 reflect more corpus-specific discourse around HBR and ILR. 

Using the macroeconomic trajectory of Topic 0 as a historical scaffold, Topic 15 and Topic 16 provide two complementary perspectives on the same broad economic transformation, one centered on corporate management and business organization, and the other on labor welfare and employment protection. In the Industrial crisis phase from 1922 to 1942, Topic 0 is centered on industrial production, trade, credit, and depression, consistent with the Great Depression and its effects on industrial production and economic activity in the 1930s \citep{federalReserveGreatDepression}. Against this macroeconomic background, Topic 15 remains concentrated on market calculation, with salient terms such as \emph{price}, \emph{profit}, \emph{sale}, and \emph{stock}. Topic 16, by contrast, highlights the labor-protection side of the same period, where \emph{hour}, \emph{wage}, \emph{insurance}, and \emph{benefit} point to employment conditions and social protection concerns under industrial crisis.

In the Corporate and welfare expansion phase from 1947 to 1977, Topic 0 shifts from crisis and production vocabulary toward firms, markets, and managerial economy. Topic 15 sharpens this shift into an explicitly managerial vocabulary: terms such as \emph{management}, \emph{manager}, \emph{executive}, and \emph{system} are consistent with the rise of large managerial enterprises and administrative coordination in postwar capitalism \citep{chandler1977visible}. Topic 16 shows the labor-side counterpart of this institutional expansion. The persistent presence of \emph{wage} is joined by terms such as \emph{union}, \emph{pension}, \emph{benefit}, and \emph{unemployment}, suggesting a stronger welfare-capitalist vocabulary around collective bargaining, employment security, and social provision \citep{espingandersen1990threeworlds}.

In the Workplace reorganization phase from 1982 to 1992, Topic 0 places more weight on work, jobs, systems, management, and services, suggesting a move toward workplace organization and service-oriented economic vocabulary. Topic 15 responds to this shift through \emph{customer}, \emph{technology}, \emph{quality}, \emph{strategy}, and \emph{process}. These terms fit broader accounts of flexible specialization, lean production, and technology-enabled organizational restructuring during the late twentieth century \citep{piore1984second,womack1990machine}. Topic 16 continues to track the labor consequences of this reorganization, suggesting that workplace change is also reflected in employment relations, gendered work, and contractual protection.

Finally, in the Digital and labor-market transformation phase from 1997 to 2017, Topic 0 turns toward internet, platform, service, customer, and innovation vocabulary, consistent with the expansion of commercial Internet services and digital-economy discourse in the 1990s and after \citep{nsfCommercialInternet}. Topic 15 translates this macroeconomic shift into a digital-era organizational vocabulary: \emph{innovation}, \emph{leader}, \emph{team}, \emph{strategy}, \emph{performance}, and \emph{organization} suggest a managerial discourse oriented toward adaptability and knowledge-intensive coordination, consistent with accounts of the networked information economy \citep{castells1996rise}. Topic 16 remains anchored in wages and employment, but terms such as \emph{minimum}, \emph{security}, \emph{care}, \emph{child}, and \emph{domestic} broaden the trajectory toward labor-market protection and social care. Taken together, the three trajectories suggest that the shared backbone learns historically meaningful base topics at different levels of economic discourse.

\subsubsection{Source-local realization of Topic 0.}
We focus the detailed source-local analysis on Topic 0: Macroeconomic transformation because its shared historical axis is broadly interpretable across COHA, HBR, and ILR, providing a clear basis for examining corpus-specific lexical differentiation. Table~\ref{tab:topic0_source_local} shows that COHA has relatively sparse and generic source-local vocabulary for Topic 0, consistent with its higher overlap with the shared backbone. HBR realizes Topic 0 through business and market-facing vocabulary, including \emph{bank}, \emph{export}, \emph{price}, \emph{product}, \emph{stock}, \emph{revenue}, \emph{share}, and \emph{customer}. ILR realizes the same aligned topic through labor-economic vocabulary, including \emph{wage}, \emph{factory}, \emph{worker}, \emph{labor}, \emph{policy}, \emph{social}, and \emph{pay}. 

We manually chose words from the learned Topic 0 with high probability in the shared or source-specific distributions at historically relevant time bins. 
We chose these words because they help illustrate both diachronic change and corpus-specific lexical realization. 
Figure~\ref{fig:raw_beta_topic0} plots the source-specific topic-word probabilities for \emph{depression}, \emph{internet}, \emph{consumer}, \emph{wage}, and \emph{labor}. \emph{Depression} receives higher probability in the early part of the trajectory and peaks around the 1930s across all three corpora, consistent with the Great Depression and its effects on industrial production and economic activity \citep{federalReserveGreatDepression}. 
This shared peak suggests that the aligned topic captures a common historical economic signal across corpora. 
\emph{Internet} increases sharply in the later part of the trajectory across all three corpora, consistent with the commercialization and rapid expansion of Internet services in the 1990s \citep{nsfCommercialInternet}. 

The remaining words show how this shared topic is realized differently across corpora. 
\emph{Consumer} captures market-facing economic discourse and is especially relevant for HBR. This pattern is consistent with the role of consumer demand and consumer markets in postwar and late twentieth-century economic discourse \citep{cohen2003consumers,americanExperienceConsumerism}. 
\emph{Wage} and \emph{labor} capture the labor-economic side of the same topic and are especially relevant for ILR, whose source-local words include labor-economic terms such as \emph{wage}, \emph{worker}, \emph{labor}, \emph{policy}, \emph{social}, and \emph{pay}. 
This pattern is consistent with ILR's focus on labor conditions and social-economic regulation \citep{internationalLabourReview,iloWages}. 
Overall, the trajectories show that the same aligned topic can support corpus-specific lexical emphasis, with each corpus assigning higher probability to words that reflect its own domain focus.
\section{Conclusion and Future Work}
We introduced a shared-backbone D-ETM framework for over-time cross-corpus topic analysis. The framework first learns a common dynamic topic backbone over a merged multi-corpus collection, then models each corpus through source-specific residual adaptation around that backbone.  Across three temporally structured corpora spanning 97 years, residual adaptation improves corpus-specific fit relative to the unadapted shared backbone while largely preserving the shared topic coordinate system. Compared with full fine-tuning from the same backbone, SB-RA achieves substantially stronger same-index trajectory alignment, with \(97.5 \pm 0.7\%\)  trajectory Retrieval@1 versus \(17.9 \pm 1.1\%\) . The ablation results further show that both residual regularizers contribute to preserving same-index topic trajectories. Qualitative analysis illustrates that the learned backbone supports historically meaningful topic trajectories and corpus-specific lexical differentiation across the three corpora studied here. These findings suggest that over-time cross-corpus comparison is better supported when models preserve not only coherent topics within each corpus, but also a stable semantic coordinate system across corpora and time. Future work can evaluate the robustness of the proposed framework through human assessments of topic interpretability. The framework can also be evaluated on additional corpus combinations with varying levels of domain overlap and temporal coverage. Another direction is to explore richer adaptation structures, such as adaptive residual strength or structured priors over source-specific drift. More broadly, the same modeling principle could be extended to newer topic modeling architectures, provided that they support explicit temporal trajectories and interpretable source-specific variation.

\begin{acks}
We acknowledge Adji B. Dieng, Francisco J. R. Ruiz, and David M. Blei
for their foundational work on the Dynamic Embedded Topic Model.
We are especially grateful to Francisco J. R. Ruiz for his generous
technical guidance. We thank Sasha Disko, Hanyu Li, Jennifer Zhang,
Max Terouanne, and Aneri B. Modi for their contributions to the
preparation and earlier analysis of the Harvard Business Review and
International Labour Review corpora, on which the present work builds.
We also thank Mark Davies, creator of the Corpus of Historical American
English, for developing and making this valuable corpus resource
available. Ruoxuan Li also thanks David Zhou for his encouragement and support throughout this project. Finally, we gratefully acknowledge Columbia Business School for
providing computational resources, technical support, and research
funding for this work.
\end{acks}
\section*{GenAI Usage Disclosure}
The authors used GenAI tools to assist with language editing and revision of manuscript text, formatting, experimental planning, and code development and debugging. All experimental outputs, analyses, citations, and reported results were verified by the authors.
\bibliographystyle{ACM-Reference-Format}
\bibliography{references}

\end{document}